\pdfoutput=1

\documentclass[11pt]{article}

\usepackage[]{acl}

\usepackage{url}

\usepackage{xcolor}
\usepackage{multirow}

\usepackage{amssymb}
\usepackage{mathtools}
\usepackage{amsthm}
\usepackage{times}
\usepackage{latexsym}
\usepackage{booktabs}
\usepackage{subfigure}
\usepackage{microtype}
\usepackage{amsmath}
\usepackage{hyperref}
\usepackage{soul}
\usepackage{graphicx}

\usepackage[capitalize,noabbrev]{cleveref}

\usepackage[T1]{fontenc}

\usepackage{amsmath,amsfonts,bm}

\def\eqref#1{equation~\ref{#1}}

\def\1{\bm{1}}

\def\rve{{\mathbf{e}}}

\def\rvx{{\mathbf{x}}}

\DeclareMathAlphabet{\mathsfit}{\encodingdefault}{\sfdefault}{m}{sl}
\SetMathAlphabet{\mathsfit}{bold}{\encodingdefault}{\sfdefault}{bx}{n}

\def\gL{{\mathcal{L}}}

\usepackage[utf8]{inputenc}

\usepackage{microtype}
\newcommand{\bc}{\boldsymbol{c}}

\newcommand{\bx}{\boldsymbol{x}}
\newcommand{\be}{\boldsymbol{e}}

\newcommand{\ename}{\textsc{EditPro}}
\newcommand{\levops}{\small{\texttt{INSERT}, \texttt{REPLACE}, \texttt{KEEP}, \texttt{DELETE}}\normalsize}

\title{Learning to Model Editing Processes}

\author{Machel Reid \\
	University of Tokyo \\
	\small{\texttt{machlereid@weblab.t.u-tokyo.ac.jp}} \\\And
  Graham Neubig \\
  Carnegie Mellon University \\
  \texttt{gneubig@cs.cmu.edu} \\}

\begin{document}
\maketitle
\begin{abstract}
Most existing sequence generation models produce outputs in one pass, usually left-to-right. However, this is in contrast with a more natural approach that humans use in generating content; iterative refinement and editing. Recent work has introduced edit-based models for various tasks (such as neural machine translation and text style transfer), but these generally model a \emph{single} edit step. In this work, we propose \emph{modeling editing processes}, modeling the whole process of iteratively generating sequences. We form a conceptual framework to describe the likelihood of multi-step edits, and describe neural models that can learn a generative model of sequences based on these multi-step edits. We introduce baseline results and metrics on this task, finding that modeling editing processes improves performance on a variety of axes on both our proposed task and related downstream tasks compared to previous single-step models of edits.%
\footnote{Data will be open-sourced at \url{https://github.com/machelreid/editpro}.}
\end{abstract}

\section{Introduction}
\vspace{-1mm}

Revising and editing are a central part of the the human creative workflow, with most original content (e.g. art, books, articles, source code) being developed not in a single iteration, but in many iterations with each more refined than the last. How can we model these \textit{editing processes} from inception to completion?
In this paper, we attempt to provide a first answer to this question, specifically focusing on generation of sequential data such as natural language documents or source code.

\begin{figure}[t]
	\resizebox{0.5\textwidth}{!}{\includegraphics[width=\textwidth]{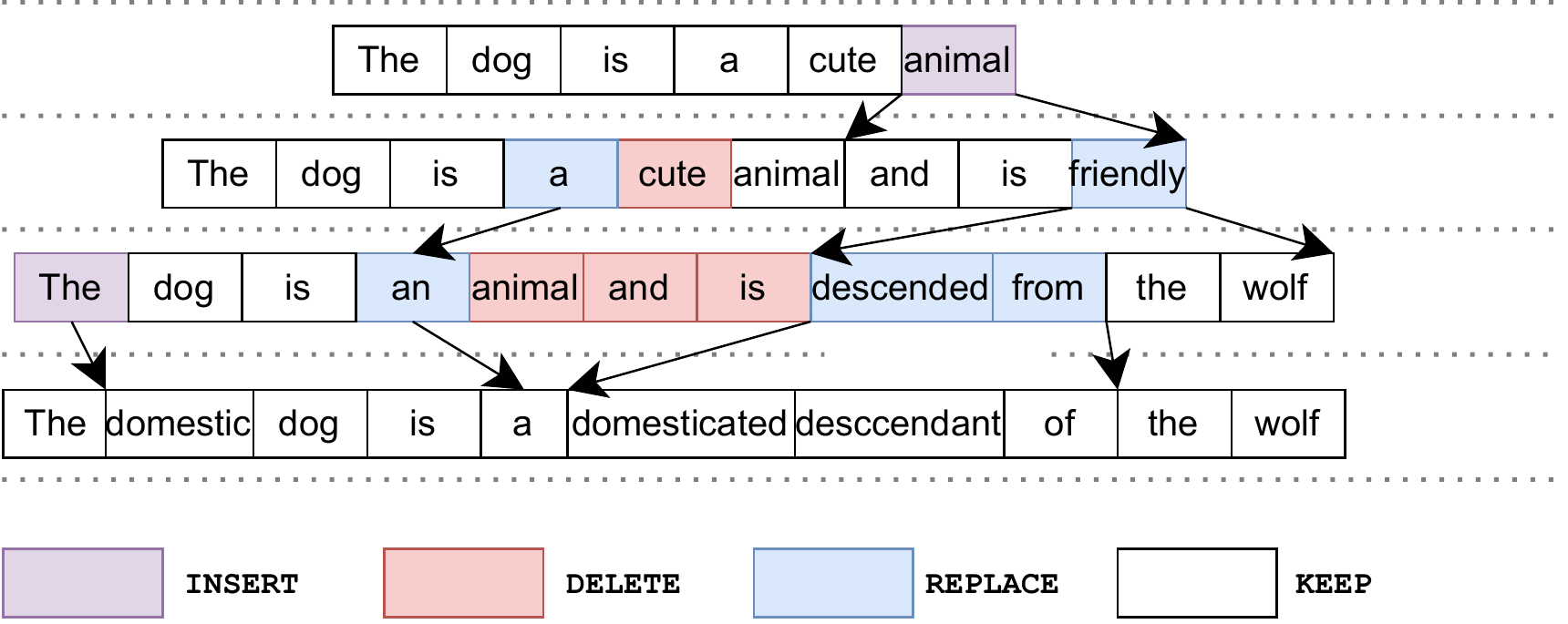}}
	\caption{An example of a natural editing process based on the description of ``Dog'' on Wikipedia. The legend below denotes the edit operations for each step of this process.}%
	\label{fig:edits}\vspace{-0.25cm}
\end{figure}

Most current work on language generation tasks such as machine translation \citep{vaswaniAttentionAllYou2017}, language modeling \citep{baevskiAdaptiveInputRepresentations2018}, or summarization \citep{seeGetPointSummarization2017} generates the target sentence or document in a single pass (usually from left to right).
There has been a reasonable amount of work that can generate edits to existing sequences for the purposes of post-editing, grammatical error correction \citep{omelianchuk2020gector}, text style transfer \citep{mallinsonFelixFlexibleText2020,malmiUnsupervisedTextStyle2020,reid2021lewis}, sentence fusion \citep{malmiEncodeTagRealize2019}, or machine translation \citep{guLevenshteinTransformer2019}. However, these works all 1) model only a single editing step and 2) do not fully define a model of incrementally editing a document from a blank slate to the final text, and thus do not stand in for the one-pass generative models of sequences described above.

In this context, we propose the task of \textit{modeling editing processes}, in which we look to explicitly model the likelihood of the entire process of revising a document to a polished form.
In particular, and in contrast to previous works on modeling edits, we hypothesize that in order to edit more accurately, instead of simply learning to predict the next revision given the current revision, we should have context of multiple previous revisions when deciding when and how to edit the document next.
Given the novelty of framing generation problems in this way, this paper simultaneously 1) proposes both baseline and novel models for the task, 2) creates evaluation datasets that can be used to compare models, and 3) discusses intrinsic and extrinsic evaluation methodology.

The proposed multi-step editing model predicts discrete edit operations \citep{levenshtein1966binary} to enable progressive refinement as shown in Figure~\ref{fig:edits}, rather than framing sequence editing as a sequence to sequence task \citep{reid2021lewis, faltings2021text}. In the figure, for each step of the editing process discrete operations (insert, replace, delete, keep) are predicted and then actions (such as generating a replaced span) are performed based on this. 
This has two benefits: 1) it allows the model to scale well with respect to input sequence length, and 2) allows us to make substantial changes with fewer actions \citep{grangier2018quickedit}. We use these edit operations to condition a semi-autoregressive model that is able to insert and replace multiple spans at once. Combined with an encoder that is able to quickly specify which spans of text need to be changed and \textit{how}, this allows for considerable changes to be made to the text (including insertion, deletion, re-ordering, and replacement) in a relatively simple and cheap manner. Furthermore, this allows us to disentangle how likely the model is to operate (replace, delete, etc.) on a given span, and how likely the model thinks the generated text for a given span is. As we are modeling editing \textit{processes}, and hypothesize that context from edits applied to the sequence are helpful, we propose a method for edit-aware sequence compression which can compress sequences into their edit operations and use \textit{relative edit positional embeddings} to specify the position of edits relative to each other.

Given that the task of modeling natural editing processes in itself is novel, we collect new datasets to study this behavior; \textsc{WikiRevisions} and \textsc{CodeRevisions}.
These datasets, in the code and natural language domains respectively, cover over 2.5M and 2.3M natural sequential revisions.
We also discuss evaluation methodology, describing a metric of \textit{edit perplexity} (ePPL), the perplexity of generating an edit given the current state of a document, as well as applications to downstream tasks.

We train and evaluate our proposed models on these datasets and find that the proposed methodology of modeling the entire editing process, referencing previous edits while generating the next one, significantly improves both intrinsic and extrinsic performance baselines that model edits in isolation. In particular, our method reduces perplexity by up to 22.9\% relative over a state-of-the-art editing baseline, and 11.3\% relative over a version of our model that does not consider editing history. We also demonstrate the ability of the model to generate qualitatively natural edit sequences, and the utility of the learned representations on downstream tasks of commit message generation \citep{loyola2017} and edit intention classification \citep{yang2017identifying}.

\section{Problem Definition}
\vspace{-1mm}
Let $X = \{\bx_0,\bx_1,\dots,\bx_N\}$ be a series of $N$ versions of a document, where the $i$th revised document is denoted by $\bx_i$.
$\bx_0$ represents an initial state (generally the null string), and $\bx_N$ represents the current state of the edited document.
The probability of this series of document versions occurring can be decomposed as
\begin{equation}
    p(X) = \prod_{i=1}^{N} p(\bx_i | \bx_0^{i-1}),
\end{equation}
where $\bx_0^{i-1} := \bx_{0}, \ldots, \bx_{i-1}$ (similarly below). 
The right hand side is the likelihood of the transformation of the previous document version $\bx_{i-1}$ to the current document version $\bx_{i}$ given the previous revision history $\bx_{<i}$.
We refer to the likelihood of the whole revision process as the \emph{edit likelihood}, and judge learned models based on their ability to achieve high edit likelihood on held-out data.

Note that standard generative models (specifically language models; LMs) calculate the probability of only the final version $p(\bx_N)$, whereas the proposed formulation calculates the probability of the entire sequence of document edits.
It nonetheless could theoretically be used to calculate the final version's likelihood by treating the editing process as latent and marginalizing over it\footnote{Or using other alternatives such as variational inference \citep{kingma2013auto}.}
\begin{equation}
    p(\bx_N) = \sum_{\tilde{X}\in\{\tilde{\bx}_1^N|\tilde{\bx}_N=\bx_N\}} p(\tilde{X}).
    \label{eq:marginal}
\end{equation}
Thus, our formulation, in contrast to previous single-step models of edits \citep{yin_learning_2019,malmiEncodeTagRealize2019,reid2021lewis}, can also be used to define a generative model over single documents.
It is also worth noting that the final document likelihood is lower-bounded by the edit likelihood; i.e.~$p(\bx_N) \ge p(X)$.

\section{Modeling Editing Processes}
\vspace{-1mm}
In this section, we now describe our approach to actually modeling these sequences of edits through (1) a decomposition of the modeling process into a sequential process of modeling edit operations then actual edits, and (2) neural model of modeling these operations and edits.
\subsection{Modeling Operations and Operation-conditioned Edits}
\vspace{-1mm}

While the probability $p(\bx_i | \bx_0^{i-1})$ of the next document given all previous document versions could theoretically be modeled with a single neural sequence model, this is infeasible computationally (and likely infeasible from learning perspective as well).
To simplify this problem, we employ the $n$-th order Markov assumption, assuming that the probability of the next document is conditioned only on the previous $n$ documents $p(\bx_i | \bx_{i-n}^{i-1})$.
This probability could be modeled directly, and in fact in the case of $n=1$ this becomes analogous to the single-step editing problem tackled by previous work \citep{yin_learning_2019,malmiEncodeTagRealize2019,reid2021lewis,faltings2021text}. To our knowledge, no previous work has modeled natural editing processes with $n>1$.

However, in the interest of both efficiency and efficacy, we take an alternative approach where we first predict a set of edit operations $\rve_i$, and then predict the next document version based on the previous documents and these edit operations:
\begin{align}
	p(\rvx_i | \rvx_{i-n}^{i-1}) & \approx p(\rvx_i, \rve_i | \rvx_{i-n}^{i-1}) \\
	& = p(\rvx_i | \rve_i, \rvx_{i-n}^{i-1}) p(\rve_i | \rvx_{i-n}^{i-1}). \label{eq:edit_prob}
\end{align}
The first approximation becomes an equality when the edit operations can be deterministically derived from $\rvx_i$ and $\rvx_{i-1}$, i.e.~$p(\rve_i | \rvx_i, \rvx_{i-1})=1$, as is the case described below.

\vspace{0.5mm} \noindent \textbf{Edit Operations.}
We base the edit operations in $\rve$ on those calculated by the Levenshtein algorithm \citep{levenshtein1966binary}, including token-level insertions, deletions, and substitutions.
These are expressed as four operations insert, delete, keep, and replace denoted by {$\{$\texttt{INSERT}, \texttt{DELETE}, \texttt{KEEP}, \texttt{REPLACE}$\}$}. For multi-word insertions and replacements, e.g. a replacement of a contiguous span of words, we apply the the same \texttt{REPLACE} label to all tokens in this span.
An example of each operation is shown in Figure~\ref{fig:edits}.

\vspace{0.5mm} \noindent \textbf{Decomposed Edit Likelihood.} We can then re-define our previous formulation of edit likelihood:
\begin{equation}
P(\rvx_1^N) = \prod_{i=1}^{N} p(\rvx_i | \rve_i, \rvx_{i-n}^{i-1}) p(\rve_i | \rvx_{i-n}^{i-1}),
\end{equation}
and analogously define edit log-likelihood %
\begin{equation}
\begin{split}
\gL_{\rvx\rve} &\coloneqq \log P(\rvx_1^N) \\
& = \sum_{i=1}^{N} \log p(\rvx_i | \rve_i, \rvx_{i-n}^{i-1}) + \log p(\rve_i | \rvx_{i-n}^{i-1}).
\label{eq:lxe}
\end{split}
\end{equation}
We can further decompose this into only the components corresponding to the edit operations
$\gL_{\rve} \coloneqq \sum_{i=1}^{N} \log p(\rve_i | \rvx_{i-n}^{i-1})$,
or the operation-conditioned edits
$\gL_{\rvx|\rve} \coloneqq \sum_{i=1}^{N} \log p(\rvx_i | \rve_i, \rvx_{i-n}^{i-1})$,
both of which we will utilize for devising evaluation metrics in Section~\ref{sec:metrics} below.

\subsection{\ename} \label{sec:model}
\vspace{-1mm}
\begin{figure*}[t]
	\centering
	\includegraphics[width=0.7\textwidth]{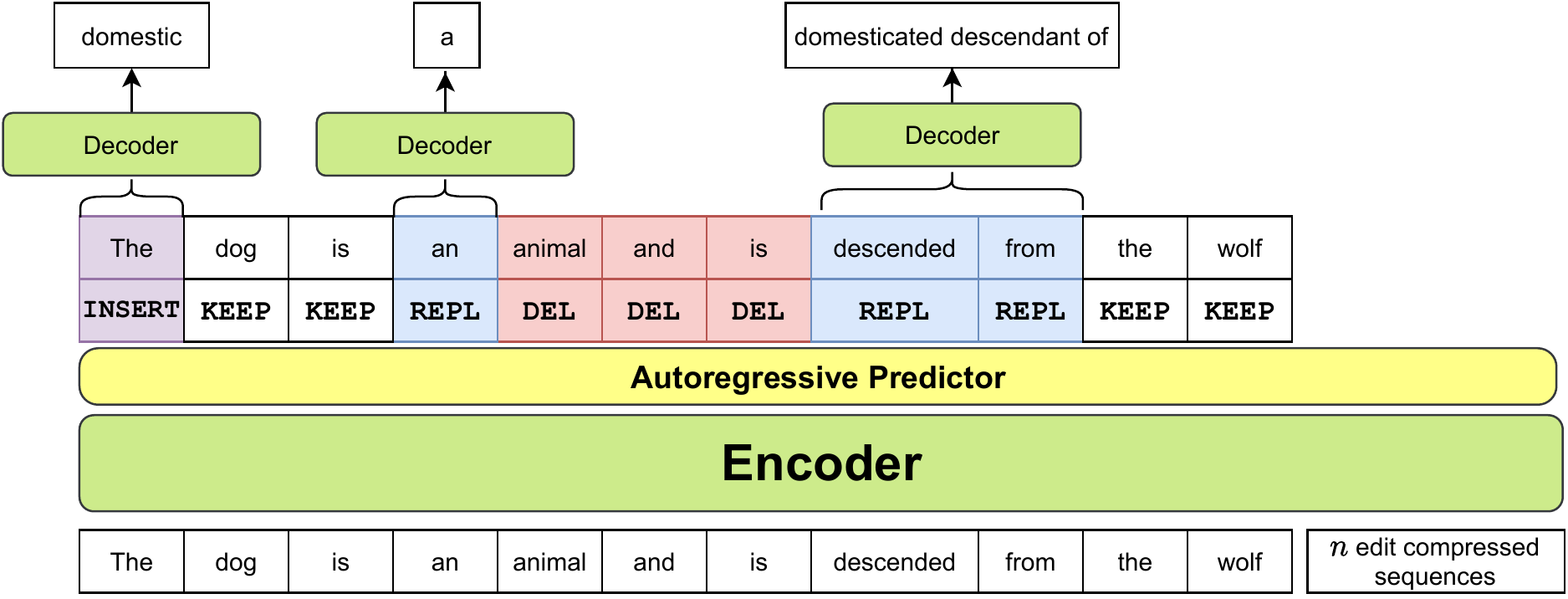}
	\caption{\ename~given the examples of modeling $p(\bx_3 | \bx_2)$ from Figure 1. We feed the input tokens into an encoder with an autoregressive tag predictor, and then use the predicted edit operations to condition the generation of \texttt{REPLACE} and \texttt{INSERT} spans.}
	\label{fig:editor}
\end{figure*}
In this section, we propose a model of multi-step editing processes, \ename, which is based on a semi-autoregressive edit-conditioned encoder-decoder model with a Transformer~\citep{vaswaniAttentionAllYou2017}. The model (depicted in Figure~\ref{fig:editor}) contains three main components: (1) an edit encoder, (2) an operation classifier and (3) an insertion-replacement decoder.

\vspace{0.5mm} \noindent \textbf{Edit Encoder.} The encoder $f_\text{enc}$ takes in a document version $\bx_{i-1}$ and feeds it through multiple self-attention and feedforward layers \citep{vaswaniAttentionAllYou2017} to produce contextual representations for each token. In the case that we perform variable-order edit modeling, we use cross-attention to feed in representations of previous edit steps.
For models where $n>1$, we feed in $n-1$ additional edit sequences -- we describe this process after describing our methods for edit sequence prediction.

\vspace{0.5mm} \noindent \textbf{Edit Operation Prediction.} We use an autoregressive tagger, using a single Transformer layer with a causal attention mask, that models the probability of each edit in edit operation sequence $\rve=e_1^M$ from left to right, $p(e_j|e_1^{j-1})$.
Notably, we also performed preliminary experiments with a tagger that predicts operations independently, but found it was heavily biased towards the \texttt{KEEP} operation as most words are kept in any single document revision, and thus did not produce coherent multi-word edit sequences when sampling sequences of edits.

\vspace{0.5mm} \noindent \textbf{Generating Replacements and Insertions.} \label{sec:decoder} When editing, given our four Levenshtein operations (\levops), two of them --- \texttt{INSERT} and \texttt{REPLACE} --- entail generation of new content conditioned on the current revision of the document.
Given our predicted edit operations $\be$, we propose a semi-autoregressive model with a causal Transformer decoder that can decode multiple spans in parallel for efficiency purposes. Each edit span contains the following properties: it has a start index (denoted by $s_\text{start}$), end index (denoted by $s_\text{end}$), and an operation type (denoted by $s_\text{type}$) .
Note that these can be simply be extracted by looking at contiguous spans of a certain type in an edit (e.g. \texttt{REPLACE} for \textit{descended from} $\rightarrow$ \textit{domesticated descendant of} in Figure~\ref{fig:edits}). 
We use a mean pooling operation to aggregate the contextual vectors produced by $f_{\text{enc}}(\bx)$ into span representation $\hat{x}_s$ : 
\begin{align}
	\hat{x}_s & = \frac{1}{s_\text{end}-s_\text{start}}\sum_{t=s_\text{start}}^{s_\text{end}}f_\text{enc}(\bx)_t
\end{align}
We then update the span representation $\hat{x}_s$ by taking the sum of the appropriate operation embedding for the span type and the current span representation and feed it to a multi-layer perceptron with an intermediate non-linearity: $\hat{x}_s \leftarrow \mathrm{MLP}(W_\text{op}(\be)_s + \hat{x}_s)$, where $W_\text{op}$ denotes an embedding matrix for each operation. $\hat{x}_s$ is then used to initialize the \texttt{<s>} token for the decoder span to further condition the generative process.

\vspace{0.5mm} \noindent \textbf{Encoding Edit History.}\label{sec:editcomp}
As we look to investigate variable order edit modeling over long sequences of text, we need a way to be able to represent edits in a way useful for predicting the next editing steps. Previous work \citep{yin2019learning,MarreseTaylor2021Variational,yao2021learning} has focused largely on learning a single-vector representation for edits which is compressed but limited in expressiveness. One the other hand, a perhaps more intuitive way taken from common Transformer-based \citep{vaswaniAttentionAllYou2017} models would be to use cross-attention between all $n$ previous documents, which is more expressive but prohibitively expensive when $n$ is scaled upwards.

Instead, we make a compromise between the above approaches, leveraging predicted edits $\rve_{i-n}^{i-1}$ to compress the sequence and their derived spans (as discussed above).
Given each of these spans, we compute the edit-compressed sequence, composed of a sequence of vector representations with each vector representing a different span. For each span in each of the previous revisions in $\bx_{i-n}^{i-1})$, we mean pool the encoder (pre-edit) and the decoder (post-edit) representations for that span. We then sum 
this representation with the operation representing its edit operation and feed it into an MLP. Once we have done this for each span, we sum a learned \textit{relative edit positional embedding}, where we learn an embedding matrix where each index in the matrix represents positions $i-1$ to $i-n$. We do this to
 specify the order of the previous edits. Finally, we compose these into a sequence and treat that as the ``edit-compressed'' sequence representation for that edit.

\vspace{0.5mm} \noindent \textbf{Turning Pre-trained Encoder-Decoder Models into Editors.}\label{sec:plm2edit}
Despite the fact that our model introduces both an edit prediction and a semi-autoregressive component, it is easy to finetune a pre-trained language model into an editor with our method as it uses vanilla Transformer layers as a backbone. We perform this by batching various spans and their conditioning variables together and training the model to adapt to decode these in parallel. 
\section{Data} \label{sec:data}
\vspace{-1mm}
While some datasets of edits exist \citep{faruqui_wikiatomicedits:_2018,MarreseTaylor2021Variational}, to our knowledge they only consider a single editing step, i.e.~dealing with a document $X = \{\bx_0,\bx_1\}, N=1$. As we propose learning to model multi-step edits, we develop new datasets in both the code and natural language domains. In addition, previous datasets have only concerned themselves with \textit{atomic} edits \citep{faruqui_wikiatomicedits:_2018} which only occur at a small scale (usually sentence-level), and we instead look to model larger-scale edits as document level changes, which are more representative of the natural editing process.

\begin{figure}[t]
	\centering
	\includegraphics[width=0.5\textwidth]{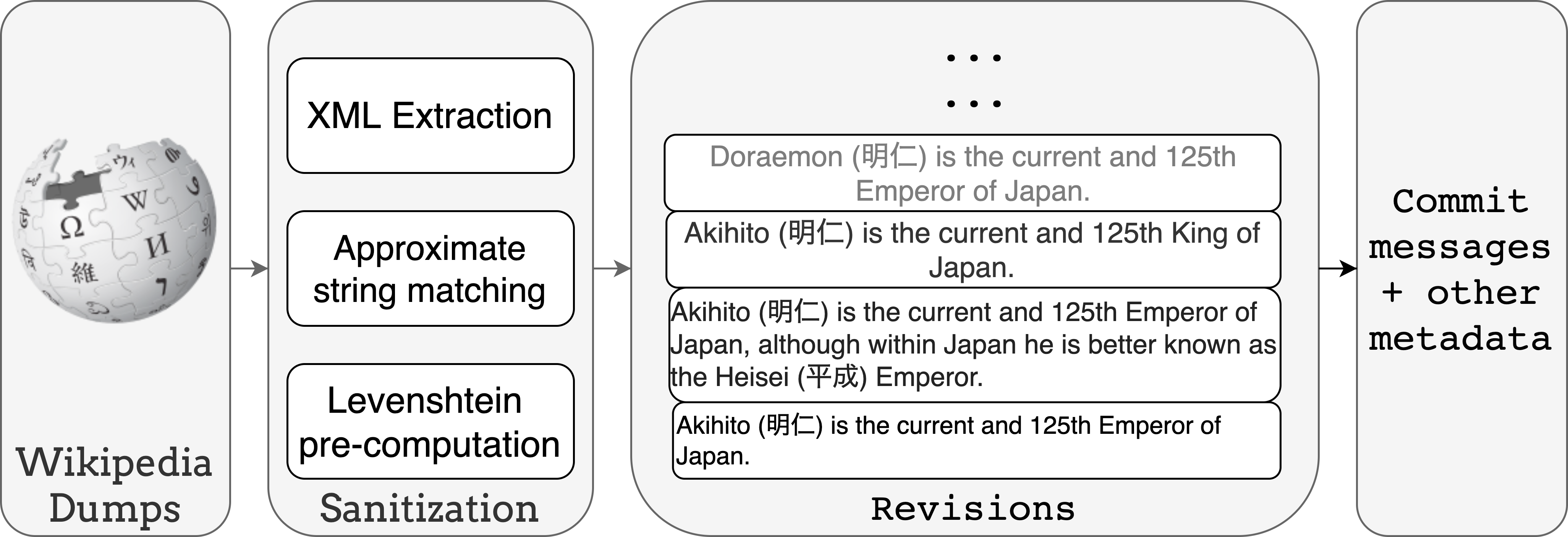}
	\caption{An overview of the \textsc{WikiRevisions} data generation process for collecting clean multi-step revision data.} %
	\vspace{-0.25cm}
\end{figure}
\begin{table*}[]
    \centering
    \footnotesize
    \begin{tabular}{lccccccccc}
	    \toprule
	    Dataset & Num. Edits & Avg. Len (Max/Min) & \% Keep & \% Insert & \% Replace & \% Delete \\
	     \midrule
	    \textsc{WikiRevisions} &  2.5M & 333 (9992/1) & 82.4\% & 0.1\% & 8.7\% & 8.8\% \\
	    \textsc{CodeRevisions} &  2.3M & 774 (9725/1) & 76.9\% & 0.1\% & 11.3\% & 11.7\% \\
	    \bottomrule
    \end{tabular}
    \caption{Dataset statistics on \textsc{CodeRevisions} and \textsc{WikiRevisions}, average length is measured by whitespace tokenization} %
    \label{tab:data}
\end{table*}
\subsection{WikiRevisions}
\vspace{-1mm}
In order to model the creative process for natural language text, we gather data from Wikipedia, which has extensive logs of the editing process that gave rise to Wikipedia articles, which have been used in a variety of previous works on single-step editing \citep{marrese-taylor_edit-centric_2019,MarreseTaylor2021Variational,yang_identifying_2017,faruqui_wikiatomicedits:_2018}.%
We collect data for each revision using dumps from English Wikipedia. Given that the dumps are provided in the XML format, we extract the text with \texttt{beautifulsoup} and remove wikitext (custom Wikipedia markup) with \texttt{wikiextractor}. With this sanitized data, we gather revision of each document in chronological order removing any metadata-based edits which were stripped as a result of the sanitization process. Now, with our sets of revisions we tokenize all text with the sentencepiece model used by \citet{radfordImprovingLanguageUnderstanding,liuRoBERTaRobustlyOptimized2019} for congruence with pre-trained models (see Section~\ref{sec:plm2edit}).
We pre-compute Levenshtein operations using 
\href{https://pypi.org/project/python-Levenshtein/}{\texttt{python-Levenshtein}}
for use during training. In the case that an article exceeds 2000 tokens, we split the articles into its subsections and treat each subsection as an article (for the purpose of modeling editing processes).
Dataset statistics are shown in Table~\ref{tab:data}. We note that there is a significant imbalance for the \texttt{INSERT} operation, this is because we define insertions to be applied to the token preceding the insertion (as shown in Figure~\ref{fig:edits}), rather than applied to an entire span (as we do for the deletion, replacement, and keep operations).

\vspace{0.5mm} \noindent \textbf{Edit Summaries.}
When extracting each edit we keep the edit summary (akin to a commit message)
supplied by the editor at time of editing. We then curate these comments and develop a dataset for usage on downstream tasks---for both edit summary generation \citep{loyola2017} and edit-summary-conditioned text editing \citep{faltings2021text}.

\subsection{CodeRevisions}
\vspace{-1mm}
Another place where the incremental creative process is on display is in the creation of program source code.
When building \textsc{CodeRevisions}, we scrape a total of 700 Python GitHub repositories using the MIT License with at least 1000 commits and 500 stars. 
We extract line-level patches from each repository's commit history when forming our code-based corpus and progressively apply each patch and compute the token-level Levenshtien operations between each revision. Note that we also keep commit messages for each commit. For this dataset we operate on the file level. For each series of revisions, we precompute Levenshtein operations based on tokens derived from a \texttt{sentencepiece} \citep{kudo2018sentencepiece} model with a 10k vocabulary. We also curate a dataset of revisions with commit messages as described in the previous subsection.

\section{Experimental Setup}
\vspace{-1mm}
\subsection{Baselines}
\vspace{-1mm}
We use the following baselines for our edit modeling task: (1) Seq2Seq, a standard sequence to sequence model trained to map $\bx_i\rightarrow\bx_{i+1}$, (2) LEWIS \citep{reid2021lewis}, a state-of-the-art
single-step editing model, which uses a separate encoder-only tagger and sequence-to-sequence generator setup during training, and (3) LaserTagger \citep{malmiEncodeTagRealize2019}, a simple editing model which learns how to apply a restricted set of edits. %
\subsection{Metrics}
\vspace{-1mm}
\label{sec:metrics}

Many previous works on editing have focused on \emph{conditional} language modeling tasks such as machine translation or editing source code based on commit messages \citep{malmiEncodeTagRealize2019, guLevenshteinTransformer2019, reid2021lewis}, and thus have used non-likelihood based metrics such as BLEU or F1 score. However as we look to model the standard \emph{unconditional} LM objective as shown in \autoref{eq:marginal}, we instead adopt a small twist on standard perplexity-based metrics from language modeling as our main intrinsic metrics. Note that $|\rvx|$ refers to the token count for the newly generated/inserted spans, and $|\rve|$ refers to the number of edit operations: 

\textbf{Edit Perplexity (ePPL)} is the exponent of the edit likelihood, divided by the length of both sequences, $\exp(\frac{-\gL_{\rvx\rve}}{|\rvx|+|\rve|})$.%
\footnote{Note that we could also divide by the length of the entire output sequence, but given that we are handling very long sequences with few changes, this would result in very small perplexities that are harder to interpret. The relative ranking of each method is unaffected regardless of presentation.}

\textbf{Generation Perplexity (gPPL)} measures the likelihood of generating replaced or inserted spans when compared with the ground truth edit sequence as follows $\exp(\frac{-\gL_{\rvx|\rve}}{|\rvx|})$.

\textbf{Operation Perplexity (oPPL)} 
$\exp(\frac{-\gL_{\rve}}{|\rve|})$ is the likelihood of predicting a set of edit operations. 

\subsection{Training Setup}
\vspace{-1mm}
We train our models using the Transformer implementation using HuggingFace \citep{wolf2020transformers}. We tokenize data using SentencePiece \citep{kudo2018sentencepiece}, using the same vocabulary used in \citet{lewisBARTDenoisingSequencetoSequence2019} for natural language, and a custom 10k vocabulary for code. We use the Transformer architecture with a hidden dimension of 768, feed-forward size of 3072, and 6 layers for both the encoder and decoder. We initialize all natural language models with BART \citep{lewisBARTDenoisingSequencetoSequence2019}, and code models randomly. We set the maximum sequence length $=$2048, using a batch size of 65K tokens distributed over 8 A100 GPUs.

\subsection{Downstream tasks}
\vspace{-1mm}
In addition to assessing our proposed model's generative capacity, we assess the quality of the learned representations on downstream tasks:
\label{sec:downstream}

\vspace{0.5mm} \noindent \textbf{Conditional Editing} We also continue training using the commit messages gathered during the cleaning process as a conditioning variable, essentially reformulating our $p(\bx_{i} | \bx_{i-n}^{i-1})$ to  $p(\bx_{i} | \bx_{i-n}^{i-1}, \bc)$ to add the additional conditional variable $\bc$, which we set to be the edit summary or commit message in this setting. With our model, we append the comment to each document %
, delimiting with a separator token \texttt{</s>} as follows: \texttt{DOCUMENT </s> COMMENT}. %

\vspace{0.5mm} \noindent \textbf{Edit-conditioned Generation.} We define edit-conditioned generation to be tasks which rely on intermediate edit representations of documents to generate text describing the changes in text, similar to that proposed by \citet{loyola2017} for source-code commit message generation. As we aim to determine whether the information contained about the edit itself is more informative as we add additional context, we condition the generation solely on the edit-compressed representations of the last edit step. To accomplish this, we use a randomly initialized Transformer decoder with cross-attention on these edit-compressed representations. 

\vspace{0.5mm} \noindent \textbf{Edit-conditioned Classification.} In the natural language domain, we also test our representations on an edit-oriented classification task, namely semantic intent classification \citep{yang2017identifying}. In \citet{yang2017identifying}, they classify 5,777 Wikipedia revisions into 10 intention classes, such as ``Clarification'', ``Vandalism'', and others with each representing a different intention. We form splits of 4,601 train examples, 588 valid examples, and 588 test examples.\footnote{We contacted the authors, but they were no longer able to release the folds they used in their K-fold cross-validation setup, hence our creation of splits} Similarly to our setup for edit-conditioned generation, we also test our classifier (consisting of a self-attentive span extractor \citep{leeEndtoendNeuralCoreference2017} and a MLP) on this task.
\section{Results}
\vspace{-1mm}

\begin{table*}[t]
    \centering
    \footnotesize
    \begin{tabular}{llcccccc}
	 \toprule
	 \multirow{2}{*}{\textbf{\sc Dataset}} & \multirow{2}{*}{\bf Model} & \multirow{2}{*}{\bf ePPL} & \multirow{2}{*}{\bf gPPL}& \multicolumn{4}{c}{\bf oPPL} \\
				      &&&& \bf \tt DEL & \bf \tt KEEP & \bf \tt REPL & \bf \tt INS\\
	 \midrule
	 \multirow{4}{*}{\textsc{\textbf{Wiki}Revisions}}& LEWIS & 65.94 & 48.85 & 24.29 & \bf 1.09 & 19.49 & 507.76 \\
	 & Seq2Seq & --- & 41.81 & --- & --- & --- & --- \\
	 & LaserTagger & --- & --- & 24.12 & \bf 1.09 & 18.45 & 889.24 \\
						& \ename \ (1-order) & 57.32 & 42.43 & 25.53 & \bf 1.09 & 18.36 & 1826.21 \\
						& \ename \ (2-order) &  53.91 & 39.87 &  20.70  & 1.13 & 15.49 & 376.15 \\
						& \ename \ (3-order) & \bf 50.84 & \bf 37.66 & \bf 19.30& 1.13 & \bf 14.88 & \bf 252.14  \\\midrule
	 \multirow{3}{*}{\textsc{\textbf{Code}Revisions}}& \ename \ (1-order) & 34.22 & 28.02 & 125.21  & \bf 1.05 & 10.38  & 544.57 \\
								  & \ename \ (2-order) & 30.85 & 26.26 & 84.77 & \bf 1.05 & 9.30 & \bf 304.90 \\
								  & \ename \ (3-order) & \bf 29.47 & \bf 25.37 & \bf 75.19 & 1.06 & \bf 8.16  & 441.42 \\
	 \bottomrule
    \end{tabular}
    \caption{Results on Edit Modeling}\vspace{-0.25cm} %
    \label{tab:em}
\end{table*}

\subsection{Edit Modeling}
\vspace{-1mm}
Results on edit modeling for both \textsc{CodeRevisions} and \textsc{WikiRevisions} can be seen in Table~\ref{tab:em}, where we measure edit perplexity, operation perplexity, and generative perplexity. We first note that our model significantly outperforms LEWIS \citep{reid2021lewis} on \textsc{WikiRevisions}, by 8.6 ePPL, suggesting that our model formulation is superior at this task. We believe that this stems from the fact that our model is trained to explicitly generate the newly added spans, and because it directly connects the operation prediction and generation processes. We also remark that although Seq2Seq gPPL is slightly lower than our baseline, it tends to learn copy given the large portion of small edits and the lack of fine-grained control over edits enabled by edit operations. LaserTagger has the opposite issue: given that they select a set of ``most common'' phrases as the model was initially proposed for sentence fusion, despite the fine-grained control provided by edit operations, generative capability is limited. For \ename~we also take note that ePPL decreases when the order of context increases. In particular, we take note of the significant gain when introducing the notion of editing processes to the model in our 2-order setting (in contrast to a single edit step), with a reduction of 3.4 ePPL on natural language and 4.4 ePPL on source code. 

We also note that while the gPPL consistently decreases as the number of orders increases, oPPL does not perform as consistently. Interestingly, we find that single-order models tend to be more confident with respect to keeping tokens (the overwhelmingly dominant edit operation), while other operations (deletions, replacements and insertions) are not predicted as well. In contrast, learning higher-order editing processes almost universally decreases the oPPL for non-KEEP operations, indicating the necessity of longer context to capture these rarer edit operations.

\noindent \textbf{\textit{Likely} and \textit{Unlikely} Edits.}
We perform a qualitative analysis on a subsample 4,000 natural language edits,\footnote{A precursory examination found these to be more interpretable than code edits.} examining which edits are judged to be likely (or unlikely) and with respect to which metrics. We do this by identifying outlier values for each metric (significantly above or below the average) and further analysing these for unique properties. 

As a result, we found that many of the edits with higher oPPL were spam and vandalism-type edits, as many of the edit operations have more of a random nature.
However we notice that generative perplexity was much lower as these edits tend to be repetitive in nature with the same ngrams often being repeated for long spans.
However, we notice that, irrespective of the number of orders, when editing \textit{reverted} spam-like content, the oPPL for the \texttt{REPLACE} and \texttt{DELETE} operations are extremely low (on average 1.07 and 4.4 respectively).
The importance of variable-order modeling was particularly evident these revisions where the gPPL in the single-order setting averages at 123.90 gPPL, however when using 2-orders we are able to attain 67.83 gPPL indicating that the edit-compressed sequences provide useful context about the previous revisions. We also notice that models are able to predict insertions (2.25 \texttt{INSERT} oPPL) significantly better when they come after the end of a sentence, representative of many insertions in Wikipedia. We also notice that outside of the above settings, models with extra context generally predict more likely edits supporting the notion of modeling edit processes compared to modeling changes individually.

\subsection{Downstream Performance} 
\vspace{-1mm}

\begin{table}[t]                                                                        \centering
\footnotesize     
\resizebox{0.5\textwidth}{!}{\begin{tabular}{llcccccc}   
\toprule             
{ \sc Dataset } & {\bf Model} & {\bf BLEU} & \bf F1 & \bf ePPL ($\Delta$)\\             \midrule              
\multirow{3}{*}{\textsc{WikiRevisions}}& \ename \ (1-order) & 10.7 & 57.8 & 54.72 (-2.60)\\                  
& \ename \ (2-order)  & 11.3 & \bf 61.3 & 51.83 (-2.08)\\     
& \ename \ (3-order) & \bf 11.6 & 61.2 & \textbf{49.91} (-0.93)\\\midrule         
\multirow{3}{*}{\textsc{CodeRevisions}}& \ename \ (1-order) & 13.8 & — & 33.65 (-0.57) \\    
& \ename \ (2-order) & 14.3 & — & 30.13 (-0.72) \\        
& \ename \ (3-order) & \bf 14.5 & — & \textbf{29.08} (-0.39) \\     
\bottomrule                                              
\end{tabular}}                                 
\caption{Results on Edit Generation (BLEU), Edit Classification (measured with micro-F1), and Conditional Edit Generation (measured Edit Perplexity = ePPL). The $\Delta$ symbol refers to the change between the model's non-message conditioned version in Table~\ref{tab:em}.} %
\label{tab:downstream}  
\vspace{-3mm}
\end{table}
Results on conditional edit generation, edit classification and edit-conditioned generation can be seen in Table~\ref{tab:downstream}. The findings generally follow the edit modeling results, with additional context improving performance further giving supporting evidence to modeling editing processes. Specifically, increasing context from single-order to 3-order improves commit message generation performance by 1.9 and 0.7 BLEU for both natural language and source code respectively. We also note that ePPL decreases similarly when we add natural language conditioning to the editing process, which indicates that multi-order editing encodes fine-grained information not found in the commit message. We note that we expect further performance gains to diminish past order 3 (as we already have diminishing returns for single-order to 2-order, and 2-order to 3-order models), however, we did not perform these experiments due to GPU memory limitations.

\begin{table*}[!t]
    \centering
    \resizebox{0.8\textwidth}{!}{
    \small
    \begin{tabular}{c|p{0.8\textwidth}}
         Initial Sentence (1-order) & Europe is a continent located entirely in the Northern Hemisphere and mostly in the Eastern Hemisphere. \\ \midrule
         $\bx_2$ & Europe is a continent located entirely in the Northern Hemisphere and mostly in the Eastern Hemisphere. \textcolor{purple}{Spain is a member of the European Union.}\\  \midrule
         $\bx_3$ &  Europe is a continent located entirely in the Northern Hemisphere and mostly in the Eastern Hemisphere. \textcolor{blue}{France} is a member of the European Union. \\  \midrule
         $\bx_4$ & Europe is a continent located entirely in the Northern Hemisphere and mostly in the Eastern Hemisphere. France is \textcolor{blue}{is a lieing country in the world. It is a bunch of crap.}\\\midrule
         $\bx_5$ & Europe is a continent located entirely in the \textcolor{red}{\st{Northern}} Hemisphere and mostly in the Eastern Hemisphere. \textcolor{red}{\st{France is a lieing country in the world. It is a bunch of crap.}} 
         \textcolor{purple}{There is a type of debate of a group of people who are not considered to be a part of the United Nations.}\\
         \midrule\midrule
         Initial Sentence (2-order) & Europe is a continent located entirely in the Northern Hemisphere and mostly in the Eastern Hemisphere. \\ \midrule
         $\bx_2$ & Europe is a continent located entirely in the Northern Hemisphere and mostly in the Eastern Hemisphere. \textcolor{purple}{The Western South Eastman Islands are also located in Europe.} \\\midrule
         $\bx_3$ & Europe is \textcolor{blue}{.k.ka.j.jf.go.skxklse} \\\midrule
         $\bx_4$ & Europe is \textcolor{red}{\st{.k.ka.j.jf.go.skxklse}} \textcolor{purple}{a continent in the Northern Hemisphere. The Islands are also in Europe and they are great.}
    \end{tabular}
    }
    \vspace{-3mm}
    \caption{Example generation when sampling with an edit model. We notice that the 2nd order model is able perform a revert operation given the context fed through the edit-compressed sequence about the previous revision, whereas the 1-order model although deleting its generated spam, generates something relatively unrelated. However we note that this reversion is not exact (likely due to the information loss during edit compression). This corresponds with our observations in our qualitative study (where likelihood of reverted edits is increased in the 2+ order models).}
    \label{tab:my_label}
    \vspace{-3mm}
\end{table*}

\vspace{0.5mm} \noindent \textbf{Edit Modeling} In particular, when performing editing using an editor pre-trained on edit modeling, we note that when sampling from the autoregressive tagger it almost always predicts \texttt{KEEP} with extremely high confidence 
, given the overwhelming class majority. We instead perform a variety of posterior regularization \citep{ganchevPosteriorRegularizationStructured}, reducing the probability of the \texttt{KEEP} class by modifying the bias term in the softmax %
until the sampled edit labels to grow closer in proportion to the true distribution of edit operations (Table~\ref{tab:data}). Combined with this technique, we are able to generate more diverse edits, which we show in Table~\ref{tab:my_label}.%

\vspace{0.5mm} \noindent \textbf{Semantic Coherence} In looking at the example generations in Table~\ref{tab:my_label} we note that the generated text is not perfectly semantically coherent, despite showing topical coherence and some degree of discourse coherence. We believe this is largely due to the size of the language model we use, being trained solely on Wikipedia data (which contains a variety of minor edits including spam/vandalism). Given this, we expect improved semantic coherence upon scaling up data, data diversity and model scale. However, we note the improved context-awareness of the edit path shown by the 2-order model over the 1-order model, providing qualitative evidence for modeling editing processes and looking at different forms of document construction.

\subsection{Human Evaluation}
We additionally perform a human evaluation using 3 Amazon Mechanical Turk crowdworkers to annotate 100 samples from our edit models at inference time. We compare our LEWIS, Seq2Seq, \ename~(1-order) and \ename~(2-order) models. Annotators are initially given 100 gold examples from the training set to in order to prime them on the form of \textit{natural} edits. Annotators are then given samples to annotate on a scale of 1 to 5, where 1 corresponds to \textit{no edit} and 5 corresponds to \textit{natural edit}, where 2,3,4 represent \textit{{somewhat natural, moderately natural, almost natural}}, respectively. %
We take the average of the three annotators' scores to provide the following results: Seq2Seq (1.7), LEWIS (2.6), \ename (1-order; 2.5), \ename~(2-order; 3.2). These results support our findings in Tables~\ref{tab:em} and \ref{tab:downstream}, demonstrating the perceptible impact of increased previous context when edit modeling (from the 2-order mdel), and the tendency to copy of Seq2Seq models faced with fine-grained editing.
\section{Related Work}
\vspace{-1mm}

\vspace{0.5mm} \noindent \textbf{Learning Edit Representations.} Previous work on learning properties inherent to editing has largely focused on learning distributed representations of edits. \citet{yin2019learning} proposed this task, using an attentional sequence-to-sequence model to learn representations. \citet{MarreseTaylor2021Variational} expands upon this approach, introducing an auxiliary loss and a variational setup. More recently, \citet{yao2021learning} worked on using this approach for editing tree-structured data. However, in contrast with this work, these approaches only consider modeling single editing steps instead of the more general multi-step setting tackled here. 

\vspace{0.5mm} \noindent \textbf{Semi-Autoregressive Sequence Generation.} Work in machine translation has explored non-autoregressive methods that use an iterative generation process. This was first proposed by \citet{lee2018deterministic} using latent variables and denoising autoencoding. Iterative refinement was then explored with conditional masked language models (CMLM; \citealp{ghazvininejad2019maskpredict}), simplifying previously proposed methods, by iteratively replacing predicted tokens with low confidence. \citet{guLevenshteinTransformer2019}, introduced the Levenshtein Transformer, making this approach more flexible by introducing insertion and deletion operations. However, these methods have not yet considered modeling \textit{natural} editing processes, instead using either random sampling or heuristically determined orders.

\vspace{0.5mm} \noindent \textbf{Other Editing-based Work.}  Other work on editing has included editing for sentence fusion \citep{malmiEncodeTagRealize2019}, in which one can perform minimal operations to join two sentences together grammatically. Furthermore, with text style transfer in which the difference between sentences in different styles (e.g. positive vs negative) can be relatively minimal \citep{reid2021lewis,mallinsonFelixFlexibleText2020}. Furthermore, \citet{faltings2021text} explored natural language conditioned editing as a means for controllable text generation using a T5-based \citep{raffelExploringLimitsTransfer2020} sequence to sequence model. Also related to our work is text morphing \citep{huang2018text}, in which they look at an edit-based model to interpolate between two sentences. We also note that diffusion models \citep{sohldickstein2015deep,ho2020denoising} can be formulated as a flavor of editing models, where the model learns to iteratively edit some representation of information in order to construct a final version of said representation. Editing-based lexically-constrained generation has been explored by \citep{miao2019cgmh} and \citet{li2020unsupervised} propose a search-based method for improving conditional text generation by way of heuristics --- an approach similar in spirit to editing, however often directed towards a specific task (which benefits from specific constraints), rather than inherently modeling sequential editing processes.
\section{Conclusions}
\vspace{-1mm}
In this work, we proposed the novel task of modeling editing processes, in which we model the likelihood of documents by way of their natural editing processes. We develop new datasets and curate existing datasets for downstream tasks. We find that modeling editing processes is beneficial to this end, in contrast to modeling single-order edits, as has been done in much of previous literature. More broadly, we believe that tackling iterative refinement tasks in broader NLP may be beneficial given its resemblance to the natural generative and creative process.

In future work, we look to investigate methods for transferring pre-trained edit models to a wider range of NLP tasks.
For instance, there are other widely acknowledged editing tasks in NLP such as grammatical error correction \citep{dahlmeier-ng-2012-better} or style transfer \citep{shen2017style}.
The technical challenge in applying the models presented here to these tasks lies in the lack of datasets annotated with multi-step edits, and potential methods to overcome this challenge would be creative use of existing datasets (e.g.~identifying relevant edits in \textsc{WikiRevisions}), or latent variable learning methods to approximate the marginal in \autoref{eq:marginal} such as variational auto-encoders \citep{kingma2013auto}.

\bibliography{custom}

\appendix
\section{Appendix}
\subsection{Further training details} \label{app:ftd}
When training, we employ data sharding to enable cheaper, on the fly data processing. We shard each documents' into 10 shards and form splits based on these shards. Our train-valid-test splits are split 90\%,5\%,5\% for commit message generation, commit-conditioned edit modeling, and edit modeling. We use a dropout value of $0.3$ and use the GELU activation for all MLPs. We use a learning rate of 1e-4 warmed up for 1000 iterations. We also note that we extended the positional embedding matrix for BART to handle longer sequences.

\end{document}